\documentclass{article} %
\usepackage[preprint]{colm2025_conference}

\usepackage{microtype}
\usepackage{hyperref}
\usepackage{url}
\usepackage{booktabs}
\usepackage{graphicx}
\usepackage{amsmath}
\usepackage{amssymb}
\usepackage{soul,comment}
\usepackage{xcolor}
\usepackage{xspace}
\usepackage{longtable}
\usepackage{array}
\usepackage{mdframed}

\usepackage{lineno}

\usepackage{algorithm}
\usepackage{algpseudocode}

\definecolor{darkblue}{rgb}{0, 0, 0.5}
\hypersetup{colorlinks=true, citecolor=darkblue, linkcolor=darkblue, urlcolor=darkblue}
\newcommand{\method}{\mbox{$\mathop{\textsc{LARC}}\limits$}\xspace}

\newcommand{\methodmistral}{\mbox{$\mathop{\method_\mathsf{Mistral}}\limits$}\xspace}

\newcommand{\methodclaude}{\mbox{$\mathop{\method_\mathsf{Claude}}\limits$}\xspace}

\newcommand{\evaluator}{\mbox{$\mathop{\textsc{Evaluator}}\limits$}\xspace}

\newcommand{\synthesizer}{\mbox{$\mathop{\textsc{Synthesizer}}\limits$}\xspace}

\newcommand{\meea}{\mbox{$\mathop{\textsc{MEEA$^{\!*}$}}\limits$}\xspace}

\newcommand{\tango}{\mbox{$\mathop{\textsc{Tango$^{*}$}}\limits$}\xspace}
\newcommand{\desp}{\mbox{$\mathop{\textsc{DESP}}\limits$}\xspace}

\newcommand{\valuefn}{\mbox{$\mathop{V}\limits$}\xspace}

\newcommand{\score}{\mbox{$\mathop{S}\limits$}\xspace}

\newcommand{\reaction}{\mbox{$\mathop{r}\limits$}\xspace}

\newcommand{\ancestors}{\mbox{$\mathop{R}\limits$}\xspace}

\newcommand{\molecule}{\mbox{$\mathop{m}\limits$}\xspace}

\newcommand{\mistral}{\mbox{$\mathop{\textsc{Mistral Nemo}}\limits$}\xspace}

\newcommand{\claude}{\mbox{$\mathop{\textsc{Claude 3.5 Sonnet}}\limits$}\xspace}

\newcommand{\gpt}{\mbox{$\mathop{\textsc{GPT-4o}}\limits$}\xspace}

\newcommand{\deepseek}{\mbox{$\mathop{\textsc{DeepSeek R1}}\limits$}\xspace}

\newcommand{\chemcrow}{\mbox{$\mathop{\textsc{ChemCrow}}\limits$}\xspace}

\newcommand{\coscientist}{\mbox{$\mathop{\textsc{CoScientist}}\limits$}\xspace}

\newcommand{\cactus}{\mbox{$\mathop{\textsc{CACTUS}}\limits$}\xspace}

\newcommand{\chemtoolagent}{\mbox{$\mathop{\textsc{ChemToolAgent}}\limits$}\xspace}

\newcommand{\liddia}{\mbox{$\mathop{\textsc{LIDDiA}}\limits$}\xspace}

\newcommand{\astar}{\mbox{$\mathop{\text{A}\!^*}\limits$}\xspace}

\newcommand{\admetai}{\mbox{$\mathop{\textsc{ADMET-AI}}\limits$}\xspace}

\newcommand{\Expert}{\mbox{$\mathop{\textsc{Expert}}\limits$}\xspace}

\newmdenv[
  backgroundcolor=gray!5,
  linecolor=gray!50,
  skipabove=6pt,
  skipbelow=6pt,
  innertopmargin=6pt,
  innerbottommargin=6pt,
  innerleftmargin=6pt,
  innerrightmargin=6pt
]{promptbox}

\newcommand{\fb}[1]{\textcolor{black}{#1}}

\definecolor{darkblue}{rgb}{0, 0, 0.5}
\hypersetup{colorlinks=true, citecolor=darkblue, linkcolor=darkblue, urlcolor=darkblue}

\title{\method: Towards Human-level Constrained Retrosynthesis Planning through an Agentic Framework}

\author{Frazier N. Baker$^1$, Daniel Adu-Ampratwum$^2$, Reza Averly$^1$, Botao Yu$^1$, \\
\textbf{Huan Sun$^1$ \& Xia Ning$^{1,2,3,4}$} \\
$^1$Department of Computer Science and Engineering, \\
$^2$Division of Medicinal Chemistry \& Pharmacognosy, \\
$^3$Department of Biomedical Informatics, \\
$^4$Translational Data Analytics Institute, \\
The Ohio State University, 
Columbus, OH, USA \\
\texttt{\{baker.3239,adu-ampratwum.1,averly.1,yu.3737,sun.397,ning.104\}@osu.edu}
}

\begin{document}

\ifcolmsubmission
\linenumbers
\fi

\maketitle

\begin{abstract}
Large language model (LLM) agent evaluators
leverage specialized tools to ground the rational decision-making of LLMs,
making them well-suited to aid in scientific discoveries,
such as constrained retrosynthesis planning.
Constrained retrosynthesis planning is
an essential, yet challenging, process within
chemistry for identifying synthetic routes from commercially available starting materials to
desired target molecules, subject to practical constraints.
Here, we present \method, the first LLM-based Agentic framework for Retrosynthesis planning
under Constraints.
\method incorporates agentic constraint evaluation, through an Agent-as-a-Judge,
directly into the retrosynthesis planning process, using agentic feedback grounded
in tool-based reasoning to guide and constrain route generation.
We rigorously evaluate \method on a carefully curated set of 48 constrained retrosynthesis
planning tasks across 3 constraint types.
\method achieves a 72.9\% success rate
on these tasks, vastly outperforming LLM baselines and approaching human expert-level success in substantially less time.
The \method framework is extensible, and serves as a first step towards an effective agentic tool or a co-scientist to
human experts for constrained retrosynthesis.
\end{abstract}

\section{Introduction}

Large language model (LLM) agents have recently shown great promise as evaluators ~\citep{zhuge_agent_2024}.
An LLM agent evaluator, also called an Agent-as-a-Judge, provides grounded evaluations in complex settings,
eliciting the strengths of general-purpose LLMs and domain-specific tools.
These evaluators are poised to make substantial impacts in scientific discoveries, such as in chemistry, 
where complex evaluation settings abound~\citep{butters_route_2011, blass_basic_2021},
accurate evaluations are paramount,
and many domain-specific tools exist to ground evaluations~\citep{ m_bran_augmenting_2024, averly_liddia_2025}.
One such application is \textit{constrained retrosynthesis planning}~\citep{corey_logic_1989},
an essential process 
in
chemistry for identifying synthetic routes from commercially available starting materials to desired target molecules (products), subject to practical constraints (e.g., avoiding hazardous 
molecules). %

LLM agents for constrained retrosynthesis planning are not explored. 
Current artificial intelligence (AI) methods are primarily focused 
on unconstrained retrosynthesis planning~\citep{chen2020retro, zhao_meea_2024, kim_self-improved_2021, genheden_aizynthfinder_2020, chen_g2retro_2023, baker_rlsync_2024, current_differ_2025},
aiming to generate 
synthetic routes that are feasible. 
Only a few AI methods have attempted to address constrained retrosynthesis planning~\citep{yu_double-ended_2024, guo_it_2024}. 
However, they only support a very simplistic constraint -- including a user-specified molecule in the synthetic routes, 
and cannot be applied to more general and practical constraints, such as avoiding broad classes of 
hazardous molecules in the synthetic routes. 
These constraints are substantially more challenging to evaluate and enforce,
requiring specialized knowledge of hazardous materials.
Rather than guiding synthesis planning towards a single, clearly-specified goal,
these constraints require guiding it away from many diverse hazards.
LLM agents are well-suited for such a challenge,
as they can leverage specialized chemistry tools to ground their evaluations and make rational decisions to guide the planning.
Furthermore, LLM agents
could support
a variety of constraints, choosing the appropriate tools for each type of constraint.
They have the potential to mimic the typical behaviors of human chemists, such as using reference materials~\citep{unece_globally_2023, iarc} to assess safety constraints during retrosynthesis planning~\citep{butters_route_2011}, and thus, 
automate, accelerate, and optimize the reliability of outcomes in the constrained retrosynthesis planning process. 

\begin{figure*}[!t]
\centering
\vspace{-5pt}
\includegraphics[width=0.99\linewidth]{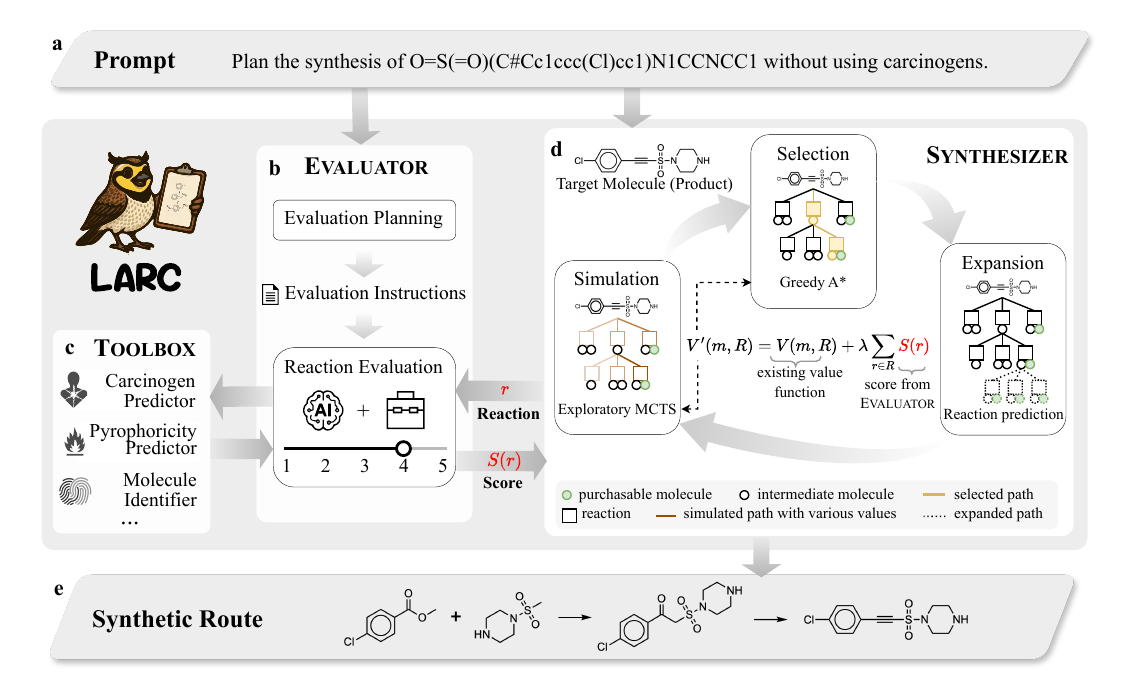}
    \vspace{-15pt}
\caption{Overview of \method. 
\textbf{a,} The user prompt specifies the target molecule (product) and the constraint for synthesis.
\textbf{b,} \evaluator acts as a judge and evaluates each individual reaction involved in the retrosynthesis planning with respect to the constraint. 
\textbf{c,} The toolbox contains external tools to ground \evaluator's decision-making
\textbf{d,} \synthesizer explores and constructs synthetic routes, incorporating feedback from \evaluator. \textbf{e,} \method outputs a synthetic route that satisfies the constraint. }
\label{fig:overview}
    \vspace{-10pt}

\end{figure*}

Here, we present \method, an \textbf{L}LM-based \textbf{A}gentic framework for \textbf{R}etrosynthesis planning under \textbf{C}onstraints. Figure~\ref{fig:overview} presents an overview of \method. 
\method uses an Agent-as-a-Judge, equipped with chemistry tools, to evaluate constraints during retrosynthesis planning.
This agentic feedback is incorporated back into the retrosynthesis planning process, dynamically guiding and constraining route generation.
It addresses key safety constraints in retrosynthesis planning, 
such as avoiding carcinogens, pyrophoric substances, or a user-specified substance.
\method is extensible by design, allowing it to improve or expand as future capabilities emerge.
To the best of our knowledge, 
\method is the first agentic framework for constrained retrosynthesis planning, 
representing an innovative paradigm for this complex scientific problem.

We rigorously evaluate \method on a carefully curated set of 48 constrained retrosynthesis planning tasks across 3 constraint types.
\method achieves an impressive 72.9\%  success rate on these tasks, indicating that \method is very effective at constrained retrosynthesis planning.
We compare \method against general-purpose LLMs and a human expert. 
The experiments show that \method vastly outperforms the LLMs and approaches expert-level success in substantially less time.
Case studies indicate that \method can mimic human expert's retrosynthesis planning logic 
and even produce better synthetic routes on some tasks. 
Further analysis reveals the key impact of agentic tooling in \method, 
enabling high success rate and efficiency through deliberate and grounded evaluations.
With further extension to cover comprehensive practical constraints, 
the \method framework can serve as an effective agentic tool or a co-scientist  
to human experts for constrained retrosynthesis.
The code and data for \method are publicly available at \url{https://github.com/ninglab/LARC}.

\vspace{-5pt}
\section{Related Work}
\label{sec:related}
\vspace{-5pt}

Recently, AI methods have emerged for constrained retrosynthesis planning.
{\tango}~\citep{guo_it_2024} and {\desp}~\citep{yu_double-ended_2024} both constrain retrosynthesis planning to include a user-specified molecule in the synthetic routes.
\tango adapts an unconstrained retrosynthesis planner to constrained retrosynthesis planning, incorporating molecule similarity to the user-specified molecule as constraint guidance.
\desp performs a double-ended search, expanding the synthetic route from the target molecule and the user-specified molecule until the route is
connected and
complete.
While \tango and \desp show that AI can perform constrained retrosynthesis planning,
they do not support more practical constraints, such as avoiding hazardous molecules.
Recently, \cite{bran_chemical_2025} introduced 
LLM-based re-ranking of synthetic routes generated by an unconstrained planner to identify those 
satisfying the constraints.
This approach is computationally expensive and assumes that some generated synthetic routes will satisfy the constraint, which may not always occur.
Furthermore, it relies on an LLM's intrinsic knowledge alone to evaluate routes,
which may be insufficient for some constraints.
Thus, there remains a need for a framework that directly incorporates agentic constraint evaluation into the planning process, enabling synthetic route generation under practical constraints.
Additional related work on LLMs for chemistry is presented in Appendix~\ref{app:related}.

\vspace{-5pt}
\section{\method: An Agentic Framework for Constrained Retrosynthesis}
\vspace{-5pt}

\method is an agentic framework for constrained retrosynthesis planning.
For a given target molecule (i.e., product), 
it effectively plans its synthetic routes that satisfy constraints specified by user prompts. 
\method incorporates constraint evaluation directly into its planning process, leveraging agentic feedback grounded in tool-based reasoning 
to dynamically guide and constrain route generation.
\method features two key components: 
\textbf{(1)} \evaluator, %
which acts as a judge and 
evaluates each individual reaction involved in the retrosynthesis planning with respect to the constraint, 
and
\textbf{(2)} \synthesizer, which explores and constructs synthetic routes, incorporating feedback from \evaluator.
Through these components, \method integrates and 
elicits the strengths of LLMs, cheminformatics tools, and 
search and planning algorithms, 
representing an innovative, extensible, agentic paradigm, with \mbox{Agent-as-a-Judge} in the loop,  
for computer-aided constrained retrosynthesis planning.

\subsection{\evaluator}

\evaluator, an LLM-based Agent-as-a-Judge~\citep{zhuge_agent_2024}, 
assesses whether reactions satisfy the constraints specified in the user prompts.
\evaluator approaches this task in two phases:
\textbf{(1)} evaluation planning, where it generates evaluation instructions specific to user constraints,
and
\textbf{(2)} reaction evaluation, where, following the evaluation instructions, 
it evaluates reactions and provides feedback to \synthesizer. 

\paragraph{Evaluation planning}
Evaluation planning occurs once for each user prompt at the beginning, 
during which \evaluator details its plan for reaction evaluation, 
including tools it plans to use and its scoring strategy, into concrete and structured evaluation instructions. 
These instructions will be consistently followed by \evaluator throughout the entire retrosynthesis planning process. 
The system instructions for evaluation planning are provided in Appendix~\ref{app:prompts}.
They were designed to encourage strategic, step-by-step evaluation %
while providing key information on tool syntax and semantics, including input/output formats.

\paragraph{Reaction evaluation} 
Following the evaluation instructions, \evaluator evaluates reactions, 
assessing how well they satisfy the constraint. 
In doing so, \evaluator adaptively switches between two action modes: 
\textbf{(1)} leveraging its own intrinsic knowledge, acting as an AI expert itself, or 
\textbf{(2)} consulting external tools in the toolbox, 
inspired by the typical behaviors of synthetic chemists 
using reference materials~\citep{unece_globally_2023, reaxys} during retrosynthesis planning.
\evaluator calibrates the evaluation outputs into scores, which will guide \synthesizer.
These scores, denoted as $\score(\reaction)$, where $\reaction$ is the evaluated reaction, 
quantify the degree of constraint satisfaction on a discrete scale from 1 to 5, 
with 1 indicating complete violation and 5 indicating full satisfaction.
\evaluator is extensible by design to incorporate new tools or adapt to tool updates, 
allowing it to improve and expand as future capabilities emerge.

\subsection{\synthesizer}

\method uses \synthesizer to generate synthetic routes for target molecules.
\synthesizer is built upon existing unconstrained retrosynthesis planners,
adapting them to constrained retrosynthesis planning.
Unconstrained planners typically leverage a search algorithm, such as A*~\citep{hart_formal_1968} 
or Monte Carlo Tree Search (MCTS)~\citep{coulom_efficient_2006},
to iterate backwards from the target molecule and search for intermediates. %
The search is expanded through single-step retrosynthesis planning 
on the current initial intermediates 
until a full synthetic route is found, starting from commercially available materials. 
The search expansion is guided by some value function, $\valuefn(\molecule, \ancestors)$, 
which estimates the utility of expanding the search along route $\ancestors$ 
upstream from its initial intermediate \molecule. 
$\valuefn(\molecule, \ancestors)$ is independently pre-trained with respect to general objectives in 
unconstrained retrosynthesis planning, such as preference for short routes with 
feasible and chemically plausible reactions.

To constrain the search, \synthesizer combines $\valuefn(\molecule, \ancestors)$ with \evaluator's score $\score(\reaction)$, 
thus producing a new, \emph{constraint-aware} 
value function $\valuefn^\prime(\molecule, \ancestors)$ to guide the search: 
\begin{equation}
\label{eqn:value}
\valuefn^\prime(\molecule, \ancestors) = \valuefn(\molecule, \ancestors) + \lambda \underbrace{\sum\nolimits_{\scriptsize{\reaction \in \ancestors}} \score(r)}_{\clap{\text{\small{constraint evaluation by \evaluator over all the reactions along the route \ancestors}}}},
\end{equation}
where
$\molecule$ is the intermediate molecule,
$\ancestors$ is a synthetic route starting from \molecule,
\reaction is a reaction in \ancestors, 
$\score(\reaction)$ is the score from \evaluator,
and $\lambda>0$ is a trade-off hyperparameter.
Thus, $\valuefn^\prime$ incorporates the constraint evaluation by \evaluator on all the reactions 
along \ancestors to guide retrosynthesis planning, ensuring the entire route is 
maximally subject to the constraint. 
This represents an innovation, differentiating \synthesizer from existing work on 
constrained retrosynthesis~\citep{yu_double-ended_2024,guo_it_2024}. 
Note that \synthesizer can easily adapt any 
unconstrained retrosynthesis planner
without retraining the original $\valuefn(\molecule, \ancestors)$, allowing \method to accommodate future advancements in retrosynthesis planning.

\vspace{-5pt}
\section{\method Instantiation}
\vspace{-5pt}

The current implementation of \method adapts \meea~\citep{zhao_meea_2024}, a state-of-the-art method for 
unconstrained retrosynthesis planning, as \synthesizer.
\meea uses a two-step process to determine how to expand its search. 
The first step uses MCTS to simulate route planning
over the expanded partial routes, %
selecting at most $K$ candidate routes for further expansion.
In this simulation, \method uses $\valuefn^\prime_{\scriptsize{\text{MCTS}}}$ as its contraint-aware value function, 
that is, $\valuefn^\prime_{\text{MCTS}} = \valuefn_{\scriptsize{\text{MCTS}}} + \lambda {\sum\nolimits_{\scriptsize{\reaction \in \ancestors}}} \score(\reaction) $ (Equation~\ref{eqn:value}), 
where $\valuefn_{\text{MCTS}}$ includes an upper confidence bound (UCB) term~\citep{lai_asymptotically_1985}
to encourage exploration.
In this case, $\score(\reaction)$ is calculated as follows: 
For the reactions that have not been evaluated by \evaluator in previous expansions, 
an optimistic default score ($\score(\reaction) = 5$) is used to further encourage exploration; 
for those that have been evaluated, their actual evaluation score $\score(\reaction)$ is used. 
In the second step of \meea, \astar search is used to select a single route for expansion from the $K$ candidate routes.
In this step, \evaluator first evaluates all the reactions in the $K$ candidate routes. 
Then, \method uses $\valuefn^\prime_{\scriptsize{\astar}}$, 
that is, $\valuefn^\prime_{\scriptsize{\astar}} = \valuefn_{\scriptsize{\astar}} + \lambda {\sum\nolimits_{\scriptsize{\reaction \in \ancestors}}} \score(\reaction) $ (Equation~\ref{eqn:value}), to select a single route for expansion.
Implementation details are presented in Appendix~\ref{app:implementation}.

\paragraph{Benchmark dataset for constrained retrosynthesis}
We carefully curated a benchmark set of constrained retrosynthesis tasks for 
a set of target molecules (products) from the USPTO-190~\citep{chen2020retro}, 
each with a single constraint to satisfy.
In this instantiation, we considered constraints of avoiding hazardous substances 
in the synthetic routes, 
which can pose serious safety risks to chemists, equipment, and the environment.
Three types of hazardous substances are included:  
\textbf{(1)} carcinogens, which are capable of causing cancer 
based on the classification by the International Agency for Research on Cancer (IARC)~\citep{iarc}; 
\textbf{(2)} pyrophoric substances, which can ignite spontaneously upon exposure to air,
according to the U.S. Navy report on air- and water-reactive materials~\citep{gibson_jack_r_handbook_1969}.
and 
\textbf{(3)} a user-specified hazardous substance (e.g., phosgene).  
To ensure these tasks are non-trivial, we selected the tasks such that the state-of-the-art 
unconstrained retrosynthesis planning methods could generate a valid route but violate the constraint, and 
the target molecule was not used in $V(\molecule, \ancestors)$ pre-training.
In the end, 48 tasks were constructed for the benchmark set. 
Figure~\ref{fig:results}\textbf{b} presents the distribution of the three types of constraints.
Table~\ref{tab:dataset} in Appendix\fb{~\ref{app:dataset}} presents the constrained retrosynthesis planning
tasks.

\paragraph{Tools for retrosynthesis constraints}
Three specific tools are supplied in the toolbox: 
\textbf{(1)} a carcinogen predictor,
which predicts whether a given molecule is a carcinogen using the state-of-the-art \admetai~\citep{swanson_admet-ai_2024} model, 
\textbf{(2)} a pyrophoricity predictor, 
which predicts the pyrophoricity of molecules by comparing them with known pyrophoric 
substances~\citep{gibson_jack_r_handbook_1969}, 
with higher molecule similarities %
indicating higher likelihood of pyrophoricity, 
and 
\textbf{(3)} a molecule identifier,
which identifies specific hazardous molecules using their fingerprints~\citep{morgan_generation_1965, rdkit}.
Outputs from the tools will be calibrated by \evaluator into scores ($S(r)$ in Equation~\ref{eqn:value}).

\vspace{-7pt}
\section{Experimental Settings}
\vspace{-2pt}
\label{sec:exp:materials}

\paragraph{Base LLMs and Baselines}

We select \mistral~\citep{mistral_nemo_2024} and \claude~\citep{anthropic_claude_2024} 
as the base LLMs for \evaluator, resulting in \method variations denoted as \methodmistral and \methodclaude, respectively. 
\mistral is selected as a representative small, open-source LLM for its strong instruction-following capabilities. 
It allows for the evaluation of how \method performs with a cost-effective, openly available model, 
thereby 
demonstrating its practicality and accessibility. 
\claude represents a state-of-the-art, closed-source LLM that has demonstrated strong performance 
in chemistry tasks and tool use~\citep{huang_2024_chemeval, yu_chemtoolagent_2025}. 
Note that \method is not limited to these two base LLMs, 
as its design is model-agnostic and can 
incorporate other LLMs as the field 
advances.

We use general-purpose LLMs, 
including \claude~\citep{anthropic_claude_2024}, \gpt~\citep{openai_gpt-4o_2024}, \deepseek~\citep{deepseek_2025}, and \mistral~\citep{mistral_nemo_2024} as the baselines for constrained retrosynthesis planning. 
We also compare \method against a human expert in retrosynthesis planning, denoted as \Expert.
\Expert is an experienced synthetic chemist with a doctoral degree and over 17 years of experience in retrosynthesis planning.
We detail our rigorous human retrosynthesis planning protocol and our LLM baselines in Appendix~\ref{app:settings}.

\begin{figure*}[t]
\centering
\vspace{-1pt}
\includegraphics[width=\linewidth]{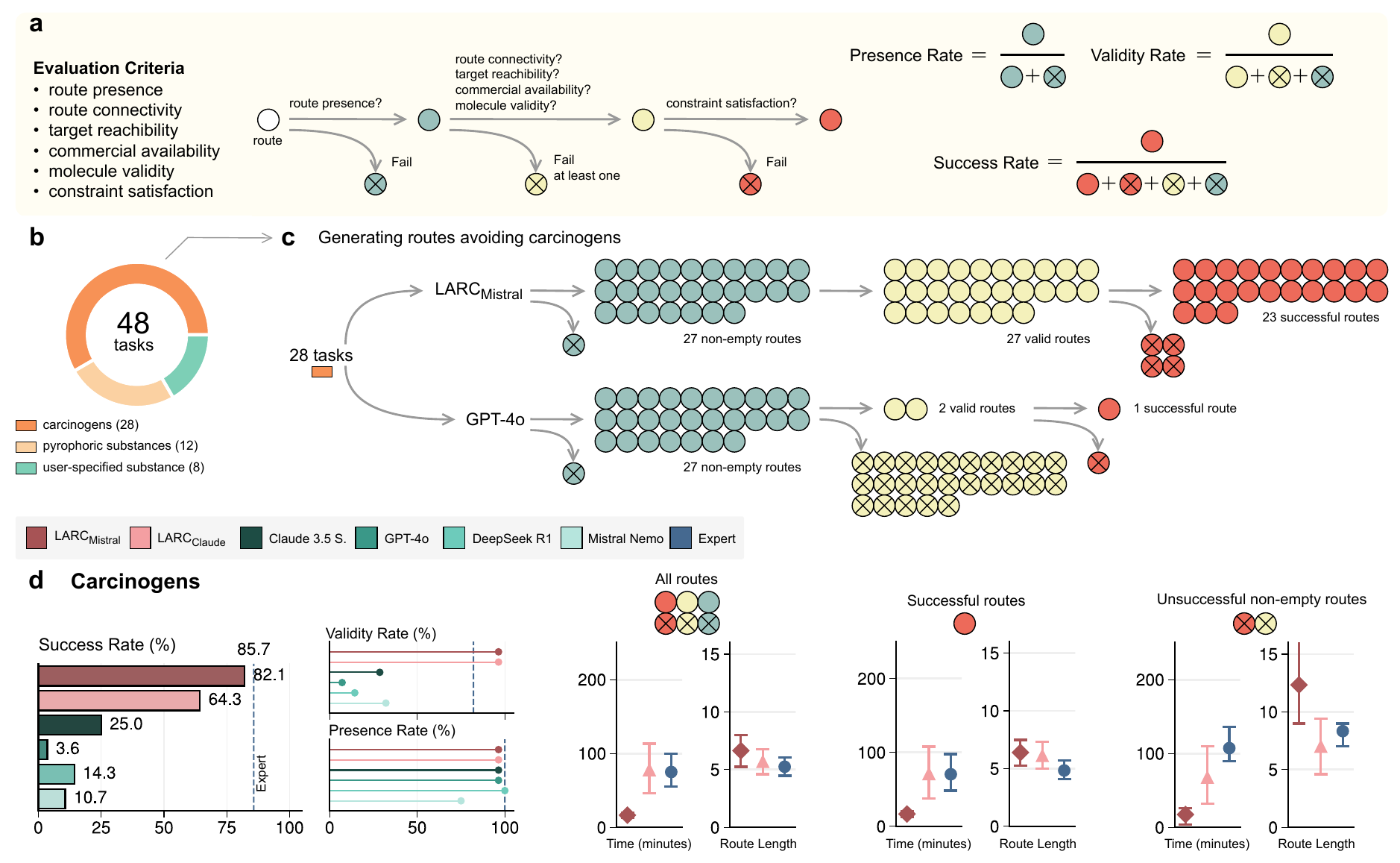}
\vspace{-15pt}

\caption{\textbf{a,} Evaluation criteria defining success rate, validity rate, and presence rate;
\textbf{b,} Benchmark dataset of 48 carefully curated constrained retrosynthesis planning tasks spanning 3 constraint types,
\textbf{c,} Example evaluation showing the calculation of presence rate, validity rate, and success rate,
\textbf{d,} success rate, validity rate, presence rate,
planning time, and route length for the benchmark tasks
for avoiding
 carcinogens.
}
\label{fig:results}
\end{figure*}

\vspace{-5pt}
\paragraph{Evaluation Metrics} The generated synthetic routes are evaluated according to the following criteria: 
\textbf{(1)} Route presence:  the routes are not empty -- they contain some molecules; 
\textbf{(2)} Route connectivity: the routes are fully connected -- 
all intermediate molecules are synthesized from preceding precursors; 
\textbf{(3)} Target reachability: the routes lead to the target molecule; 
\textbf{(4)} Commercial availability: the starting materials of the routes are directly purchasable from eMolecules dataset~\citep{chen2020retro}; 
\textbf{(5)} Molecule validity: the molecules involved in the routes are chemically correct; and
\textbf{(6)} Constraint satisfaction: the routes meet the specific constraints (avoid certain substances).  
Based on these criteria, the following metrics are used to evaluate synthetic routes, as illustrated in Figure~\ref{fig:results}\textbf{a}:
\textbf{(1)} \textbf{Success rate}: the percentage (\%) of routes that satisfy all the six criteria, that is, 
successful routes; 
\textbf{(2)} \textbf{Validity rate}: the percentage (\%) of routes that satisfy route presence, route connectivity, target reachability, commercial availability, and molecule validity, that is, valid routes per se but they do not necessarily satisfy the constraint; 
\textbf{(3)} \textbf{Presence rate}: the percentage (\%) of routes that satisfy route presence, that is, non-empty routes. 

\vspace{-5pt}
\section{Results}
\label{sec:results}
\vspace{-5pt}

We focus primarily on carcinogenicity-constrained tasks here, with results of the pyrophoricity-constrained and user-specified-constrained tasks in Appendix~\ref{app:results}.

\subsection{Carcinogenicity-constrained retrosynthesis planning}

Figure~\ref{fig:results}\textbf{d} shows that 
\methodmistral achieves a 82.1\% success rate
and \methodclaude achieves 64.3\%
in planning synthetic routes that avoid carcinogenic substances,
both vastly 
outperforming the best LLM baseline \claude (success rate only 25.0\%). 
\claude, as a general-purpose LLM, lacks specialized training on retrosynthesis planning~\citep{anthropic_claude_2024, yu_chemtoolagent_2025}, and 
fails in the vast majority of the cases. 
Unlike general-purpose LLMs, \methodmistral and \methodclaude are agentic and 
intentionally designed for this constrained retrosynthesis, 
eliciting the strengths of LLM reasoning, domain-specific tools, and efficient search algorithms.
The results demonstrate that \method is very effective in 
generating synthetic routes avoiding carcinogenic substances. %

\Expert generates 28 routes with a success rate of 85.7\%, only slightly outperforming \methodmistral.
The successful routes by \Expert have an average length of 4.83, 
which is shorter than that of \methodmistral (6.39). 
This indicates that \Expert may rely on domain-specific heuristics, intuitive shortcut strategies, 
or tacit knowledge not yet fully captured by \methodmistral.
However, \Expert requires on average 70.42 minutes to generate each successful route, 
substantially slower than \methodmistral (16.45 minutes)
(p=2.8e-4, two-sided two-sample t-test).
This highlights the potential for \methodmistral to accelerate constrained retrosynthesis planning 
while maintaining near-human-level quality.

\method and \Expert exhibit different behaviors in their unsuccessful constrained retrosynthesis planning.
\method is diligent about planning valid routes; all of its non-empty routes are valid.
However, \method's routes can sometimes violate the constraint.
This is often the result of errors in the tools outputs and their interpretation, which is discussed further in Section~\ref{sec:results:ablation} and Appendix~\ref{app:results:tool}.
In contrast, \Expert is conscientious about constraint satisfaction, but this seems to distract from route validity.
Specifically, \Expert's synthetic routes may not reach the target molecule, or it may require starting materials that are not commercially available.
This highlights a key challenge of constrained retrosynthesis planning for both \method and \Expert: balancing constraint satisfaction with validity criteria.

Interestingly, \methodmistral outperforms \methodclaude when planning routes that avoid carcinogenic substances.
\methodmistral uses the small, open-source \mistral with only 12 billion parameters as its base model~\citep{mistral_nemo_2024},
whereas \methodclaude uses the much larger, proprietary \claude with over 175 billion parameters~\citep{anthropic_claude_2024}.
\methodclaude tends to be slower -- 77.56 minutes on average to generate each route, 
than \methodmistral (16.45 minutes),
and evaluate more reactions per route (59.54) on average than \methodmistral (30.8) during the planning process.
While it could be a general expectation that larger models perform better due to scaling laws~\citep{kaplan_scaling_2020}, 
it is observed that \method enables the use of smaller and thus cheaper models, such as \mistral, without sacrificing performance
by grounding LLM reasoning with specialized chemistry tools.
Therefore, \method provides a fast, accurate, and inexpensive solution to constrained retrosynthesis planning
with low costs and a high success rate.

Aside from \claude,  the other LLM baselines also perform poorly, %
with the best success rate of only 14.3\%. %
For example, 
\gpt %
manages to generate only 1 successful route (success rate 3.6\%) among only 2 valid routes. 
This further highlights the necessity of specialized tools for constrained retrosynthesis planning -- a specific, very challenging yet important
task, and \method fills this gap.

\subsection{Case Studies}

\begin{figure*}[t]
    \centering
    \includegraphics[width=\linewidth]{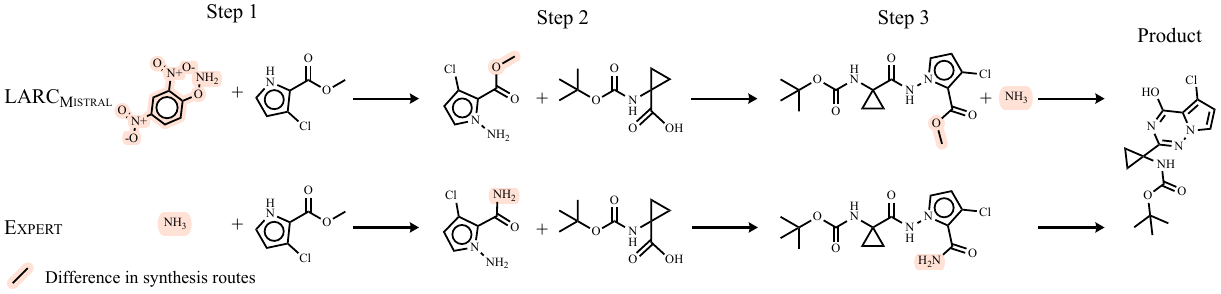}
    \vspace{-16pt}
\caption{Synthesis route comparison of \methodmistral against \Expert for tert-butyl
(1-(5-chloro-\allowbreak4-hydroxypyrrolo\mbox{[2,1-f]}\mbox{[1,2,4]}-\allowbreak\mbox{triazin-2-yl})\-cyclopropyl)\allowbreak{}carbamate.
}
\vspace{-15pt}
\label{fig:case:carc}
\end{figure*}

Figure~\ref{fig:case:carc} shows two plans for the synthesis of the product 
tert-butyl(1-(5-chloro-\allowbreak4-hydroxypyrrolo\mbox{[2,1-f]}\mbox{[1,2,4]}-\allowbreak\mbox{triazin-2-yl})\-cyclopropyl)\allowbreak{}carbamate:
one from \methodmistral and one from \Expert.
Both \methodmistral and \Expert proposed very similar three-step routes to the final product. 
In both routes, Step 1 involves the amination of commercially available methyl 3-chloro-1H-pyrrole-2-carboxylate, followed by an amidation reaction in Step 2.
Finally, Step 3 shows a condensation/cyclization sequence to provide the product. 

The main difference between the two proposed routes lies in the source of the aminating agent in Step 1.
\methodmistral proposes O-(2,4-Dinitrophenyl)hydroxylamine, to prepare the hydrazine (methyl 1-amino-3-chloro-1H-pyrrole-2-carboxylate),
which is known for being a very mild and selective aminating 
agent for introducing an NH$_2$
group into a molecule, making it a reagent of choice, especially when a metal-free process is desired.
In contrast, \Expert opted for ammonia (NH$_3$) in Step 1.
NH$_3$ is often a preferred reagent for large-scale amination reactions due to its availability and low cost \citep{verma_employing_2022}.
However, using NH$_3$ can present some challenges in activation and selectivity, often necessitating sophisticated metal catalyst systems.

As shown in Figure~\ref{fig:case:carc},
\methodmistral's use of the mild aminating agent in Step 1 will selectively introduce the NH$_2$ 
group on the pyrazole nitrogen without affecting the ester.
Conversely, the use of non-selective NH$_3$ by \Expert %
will convert the ester functional group to an amide.
Both the ester and the amide can be respectively converted to the product in Step 3, 
completing the routes.
Overall, both proposed routes have similar intermediates and transformations.
Note that \methodmistral's use of O-(2,4-Dinitrophenyl)hydroxylamine is not an anomaly---
58 patented reactions in the USPTO reaction dataset~\citep{lowe_chemical_2017} also use this reagent.
Overall, \methodmistral can mimic human retrosynthesis planning logic,
even in this challenging constrained retrosynthesis setting.
An additional case study, showing \method can even generate better routes than human experts, is presented in Appendix~\ref{app:results:case}.

\subsection{Study on tooling in \method}
\label{sec:results:ablation}

For \methodmistral, tools are critical: across all 48 benchmark tasks,
\methodmistral achieves a success rate of 72.9\% using tools in addition to \evaluator's internal knowledge,
and 45.8\% without tooling.
Specifically, 
\methodmistral with tooling generates 43 valid routes (validity rate of 89.6\%), of which 35 are successful, %
and without tooling, 41 valid routes (validity rate of 85.4\%), of which 22 are successful. %
This indicates that tooling primarily impacts \method 
through deliberately assessing and enforcing constraint satisfaction to ensure successful routes. 
Moreover, with tooling, {\methodmistral} generates successful routes faster (14.53 minutes on average) %
than without tooling (22.08 minutes), %
whereas in the latter case, {\evaluator} has to act as the AI expert and conduct reasoning via {\mistral}.  
Additional discussion on tools is available in Appendix~\ref{app:results:tool}.

\vspace{-5pt}
\section{Discussions and Conclusions}
\vspace{-5pt}

\method addresses only a very simplified version of practical constrained retrosynthesis planning, 
with the primary goal of demonstrating the potential of agentic AI in solving such complex scientific problems. 
Our experiments clearly show that, when equipped with appropriate tools for verifying constraint satisfaction, 
agentic AI can approach human expert-level performance while being more autonomous, effective, and scalable. 
Such \fb{traits}
are highly attractive in scientific workflows, as they reduce reliance 
on time-consuming and potentially inconsistent, and error-prone manual efforts, 
and allow for the exploration beyond existing, potentially outdated knowledge of human experts. 

Meanwhile, rigorous,
systematic, and scalable
evaluation and validation of results from agentic AI, including \method, still fall far short. 
For retrosynthesis, there is not always a definitive and consensus ``ground truth" for reaction feasibility.
Compounded with the inherent biases of the search space 
in unconstrained retrosynthesis planners used in \method, 
it is still likely that \method generates synthetic routes that appear 
valid but are actually infeasible. 
Even worse, these routes may be difficult to detect and filter using either tools or human expertise.
A more reliable option is to incorporate physics-based models, such as molecular dynamics simulations, 
to assess reaction feasibility in agentic AI models. 
Unfortunately, this approach requires a lot of customization (e.g., specific force fields) and 
is not readily scalable, undermining the advantages of agentic AI 
in being both autonomous and scalable.
Ultimately, testing of the AI-generated synthetic routes
in a laboratory will be needed to 
validate the results from agentic AI and truly translate its advantages into real impacts. 
This will require a selection of the most promising routes, which will eventually still rely on computational tools or 
human expertise, suffering from the same issues as in silico evaluation.  
Meanwhile, large-scale in vitro validation of in silico-generated reactions is still challenging. 
Thus, though very promising, this research on agentic AI for constrained retrosynthesis 
calls for 
in-depth and systemic investigation on its autonomous and scalable evaluation and validation strategies. 
Additional discussion is available in Appendix~\ref{app:discussion}.

\paragraph{Conclusions}
\method is the first agentic framework for constrained retrosynthesis planning, 
which uses an LLM-based Agent-as-a-Judge to evaluate reactions and guide the constrained planning. 
\method achieves a 72.9\% success rate over a carefully curated benchmark 
of 48 constrained retrosynthesis planning tasks spanning 3 constraint types. 
It vastly outperforms general-purpose LLM baselines and approaches human expert-level performance.
\method can mimic the retrosynthesis planning logic of human experts
and can even produce synthetic routes better than human experts.
These results establish \method as a compelling proof-of-concept 
for leveraging agentic AI to advance next-generation scientific discovery in synthetic chemistry. 
More broadly, it illustrates the transformative potential of agentic AI to accelerate progress across the sciences.

\vspace{-10pt}
\section*{Acknowledgments}
\vspace{-10pt}
This project was made possible, in part, by support from the National Science Foundation grant nos. IIS-2133650 
and IIS-2435819, 
and the National Library of Medicine grant no. 1R01LM014385. Any opinions, findings, conclusions or recommendations expressed in this paper are those of the authors and do not necessarily reflect the views of the funding agency.
We would like to thank our colleague, Xinyi Ling, for assisting with some of the figures.

\clearpage

\section*{Ethics Statement}

\method is an agentic framework for constrained retrosynthesis planning, generating synthetic routes for target molecules under practical constraints.
We acknowledge that not all target molecules or synthetic routes are safe, and \method could generate harmful synthetic routes.
Thus, we strongly recommend conscious oversight and intervention from human experts when using \method.
Synthetic chemists should confirm the safety of \method's generated synthetic routes before using them in laboratory experiments.
We recommend using both automated and manual safety checks against external sources (e.g. local safety policies, GHS~\citep{unece_globally_2023}, etc.),
and following standard procedures for laboratory safety.
We advise all users of \method to exercise their professional discretion and follow all applicable safety guidelines, laws, regulations, and ethical standards.

\bibliography{colm2025_conference}
\bibliographystyle{colm2025_conference}

\clearpage
\appendix
\renewcommand{\thefigure}{\thesection\arabic{figure}}

\renewcommand{\thetable}{\thesection\arabic{table}}
\section{Dataset}
\label{app:dataset}

Table~\ref{tab:dataset} shows the benchmark dataset curated for our experiments.  Tasks are provided as natural language prompts, organized by constraint type.
{
\renewcommand{\arraystretch}{1.2}
\begin{longtable}{>{\raggedright\arraybackslash}p{0.03\linewidth}
>{\raggedright\arraybackslash}p{0.15\linewidth}
>{\raggedright\arraybackslash}p{0.70\linewidth}}
\caption{Constrained Retrosynthesis Planning Dataset}\label{tab:dataset}\\
\toprule
ID & Constraint Type & Task \\
\midrule
\endfirsthead

\toprule
ID & Constraint Type & Task \\
\midrule
\endhead

\midrule
\multicolumn{3}{r}{\emph{Continued on next page}}\\
\midrule
\endfoot

\hline
\endlastfoot

C1  &  \mbox{Carcinogen}     & Find the shortest synthesis route for \texttt{O=S(=O)(C\#Cc1ccc(Cl)cc1)N1CCNCC1} that   avoids all carcinogenic substances.                                                                            \\
C2  &  \mbox{Carcinogen}     & Identify a synthesis route for   \texttt{CC(C)(C)OC(=O)NC1\allowbreak{}(c2nc(O)c3c(Cl)ccn3n2)\allowbreak{}CC1} that avoids known or suspected   carcinogens.                                                                     \\
C3  &  \mbox{Carcinogen}     & Plan the synthesis of \texttt{CC(=O)OCc1nc2cnc(Br)cc2n1C(C)(C)COC(C)=O},   avoiding any carcinogenic substances.                                                                                       \\
C4  &  \mbox{Carcinogen}     & Plan the synthesis of   \texttt{FC(F)(F)Cn1ncnc1-c1cc2n(n1)-c1cc(C3CCNCC3)\allowbreak{}ccc1OCC2}, avoiding any known or suspected carcinogens.                                                                    \\
C5  &  \mbox{Carcinogen}     & Perform synthesis planning for   \texttt{C=C(C{[}C@@H{]}(Cc1ccc(-c2ccccc2)cc1)\allowbreak{}NC(=O)OC(C)(C)C)C(=O)O} without using any   carcinogenic substances.                                                     \\
C6  &  \mbox{Carcinogen}     & Provide a synthesis route for   \texttt{COc1cc2c(Oc3cc(C)c(C)nc3-\allowbreak{}c3cccc(C)n3)ccnc2cc1OCCNCCO} that avoids   carcinogens.                                                                               \\
C7  &  \mbox{Carcinogen}     & Find the shortest synthesis route possible for   \texttt{C\#CC1(O)C(C)=CC2(CC1(C)C(F)(F)F)OC(C)C(C)O2} without using carcinogens.                                                                      \\
C8  &  \mbox{Carcinogen}     & Identify the shortest possible synthetic route for   \texttt{ClCc1ccc2c(c1)Nc1nccnc1S2} that avoids carcinogens.                                                                                       \\
C9  &  \mbox{Carcinogen}     & Identify the best synthesis route for   \texttt{COC(=O)c1ccc2c(c1)C=CC(=C(Cl)Cl)CO2} that avoids carcinogenic materials.                                                                               \\
C10 &  \mbox{Carcinogen}     & Design a synthesis plan for   \texttt{C{[}C@@H{]}(O)c1nc2cnc3ccsc3c2n1\allowbreak{}{[}C@H{]}1CC{[}C@H{]}\allowbreak{}(CO)CC1} without using any   carcinogens.                                                                   \\
C11 &  \mbox{Carcinogen}     & Provide the shortest synthesis route for   \texttt{Cn1oc(=O)nc1/C(=N\textbackslash{}\textbackslash{}OCc1cccc(N)n1)c1ccccc1} that does not use any   carcinogens.                                       \\
C12 &  \mbox{Carcinogen}     & Plan the synthesis of \texttt{CC(C)(CO)n1c(CO)nc2cnc(Br)cc21} without using   carcinogens.                                                                                                             \\
C13 &  \mbox{Carcinogen}     & Plan the synthesis of   \texttt{CCCC{[}Sn{]}(/C=C/C1(O)C(C)=CC2\allowbreak{}(CC1(C)C(F)(F)F)OC(C)C(C)O2)\allowbreak{}(CCCC)CCCC}, but do   not use any carcinogens in your synthesis route.                                      \\
C14 &  \mbox{Carcinogen}     & Find the shortest synthesis path for   \texttt{COC(=O)CCc1cc2cc(-c3noc(-c4ccc(OC(C)C)c(Cl)c4)n3)ccc2n1C} that does not use   any carcinogenic substances.                                              \\
C15 &  \mbox{Carcinogen}     & Plan the synthesis of   \texttt{CC(C)(C)OC(=O)N1CC=C(c2ccc3c(c2)-n2nc\allowbreak{}(-c4ncnn4CC(F)(F)F)cc2CCO3)CC1} without   using any carcinogens.                                                                  \\
C16 &  \mbox{Carcinogen}     & Find the shortest synthesis route for   \texttt{C{[}Si{]}(C)(C)CCOCn1cc(C2CCc3c(C(=O)O)nn(COCC{[}Si{]}(C)(C)C)\allowbreak{}c3C2)cn1} that doesn't   use any carcinogenic substances.                                \\
C17 &  \mbox{Carcinogen}     & Plan the synthesis of \texttt{O=C(OCc1ccccc1)N1CC{[}C@H{]}2CCCN\allowbreak{}(CCc3ccccc3)C{[}C@H{]}21}   without using carcinogens.                                                                                  \\
C18 &  \mbox{Carcinogen}     & Find a synthesis route for   \texttt{Cn1oc(=O)nc1/C(=N\textbackslash{}OCc1cccc\allowbreak{}(NC(=O)OCCc2ccccc2)n1)c1ccccc1} that avoids all   known or suspected carcinogens.                        \\
C19 &  \mbox{Carcinogen}     & Plan the shortest synthesis route for   \texttt{CC(C)(C)OC(=O)NC1(c2nc(NCc3ccccn3)c3c(Cl)ccn3n2)CC1}. Do not use any   carcinogens in the route.                                                       \\
C20 &  \mbox{Carcinogen}     & Generate a synthesis plan for the following compound without using any   carcinogens: \texttt{CC{[}C@@H{]}(OC(=O)c1ccccc1){[}C@H{]}1CCCN(C(=O)OC(C)(C)C)C1}                                            \\
C21 &  \mbox{Carcinogen}     & Plan the synthesis of   \texttt{CC(C)(C)OC(=O)N{[}C@@H{]}1c2cccnc2{[}C@H{]}\allowbreak{}(O)CC{[}C@H{]}1c1cccc(F)c1F}. Avoid   carcinogenic materials.                                                               \\
C22 &  \mbox{Carcinogen}     & Find the shortest synthesis route that doesn't use carcinogens for the   following product:   \texttt{CC(C)(C)OC(=O)N{[}C@@H{]}1\allowbreak{}c2cccnc2{[}C@H{]}(N)CC{[}C@H{]}1c1cccc(F)c1F}.                         \\
C23 &  \mbox{Carcinogen}     & Design a synthesis path for   \texttt{CC(C)(C)OC(=O)N{[}C@@H{]}1\allowbreak{}c2cccnc2C(=O)\allowbreak{}CC{[}C@H{]}1c1cccc(F)c1F}. Do not use any   carcinogens.                                                                  \\
C24 &  \mbox{Carcinogen}     & Plan the shortest possible synthesis route for \texttt{COCCCc1cc(CN(C(=O){[}C@H{]}2CNCC{[}C@@H{]}2c2ccc\allowbreak{}(OCCOc3c(Cl)cc(C)cc3Cl)cc2)C2CC2)\allowbreak{}cc(OCCOC)c1},   but do not use any carcinogens in the route. \\
C25 &  \mbox{Carcinogen}     & Find the shortest synthesis path for   \texttt{COCCCc1cc(CN(C(=O){[}C@H{]}2CN(C(=O)OC(C)(C)C)CC{[}C@@H{]}2\allowbreak{}c2ccc(OCCOc3c(Cl)cc(C)cc3Cl)cc2)C2CC2)cc(OCCOC)c1}.   Do not use any carcinogens.            \\
C26 &  \mbox{Carcinogen}     & Generate a synthesis plan without carcinogens for   \texttt{CC\#CCn1c(Br)nc(C=O)c1C(=O)OC}.                                                                                                            \\
C27 &  \mbox{Carcinogen}     & Find the shortest synthesis path for \texttt{COC(=O)c1ccc2c(c1)C=CC(=CCl)CO2}   that does not use carcinogenic substances.                                                                             \\
C28 &  \mbox{Carcinogen}     & Find a synthesis path for   \texttt{O=C(Nc1cccc(Cl)c1)N1CCc2{[}nH{]}nc\allowbreak{}(C(=O)N3CC(F)CO3)c2C1} that doesn't use   carcinogens.                            \\
\midrule
P1  &  \mbox{Pyrophoric}     & Identify a synthesis route for   \texttt{C{[}C@H{]}(O{[}Si{]}(C)(C)C(C)(C)C){[}C@@H{]}1CC(=O)CC(C)(C)N1} that does not use   pyrophoric substances.                                                    \\
P2  &  \mbox{Pyrophoric}     & Find the shortest synthesis plan for   \texttt{C{[}C@H{]}(c1ccccc1)N1C{[}C@{]}2(C(=O)OC(C)(C)C)C=CC{[}C@@H{]}2C1\allowbreak{}=S} that avoids all   pyrophoric and water-reactive substances.                        \\
P3  &  \mbox{Pyrophoric}     & Plan the shortest synthesis route for \texttt{CC(=O)c1ccc2c(c1)C=CC(O)(CO)CO2}   without using any pyrophoric or water reactive substances.                                                            \\
P4  &  \mbox{Pyrophoric}     & Synthesize \texttt{CC{[}C@@H{]}(OC(=O)c1ccccc1){[}C@H{]}1CCCN(C(=O)OC(C)(C)C)C1} without   using any pyrophoric or water-reactive substances.                                                          \\
P5  &  \mbox{Pyrophoric}     & Perform synthesis planning for \texttt{CC1=NC2(N=C1N)c1cc(Br)ccc1CCC21CC1},   avoiding pyrophoric materials (substances that ignite in moisture or air) in   your synthesis route.                     \\
P6  &  \mbox{Pyrophoric}     & Plan a synthesis route for   \texttt{COc1cc2ncc3c(N)nc(-c4cncc\allowbreak{}(OCCN(Cc5ccc(F)cc5)C(=O)OC(C)(C)C)c4)cc3c2cc1OC}   that uses no pyrophoric or water-reactive substances.                                 \\
P7  &  \mbox{Pyrophoric}     & Perform synthesis planning for   \texttt{O{[}C@H{]}1C{[}C@H{]}(c2cnn3c(N{[}C@H{]}4CCc5ccccc54)ncnc23)\allowbreak{}C=C1COCc1ccccc1}.    using pyrophoric substances in your synthesis plan.                     \\
P8  &  \mbox{Pyrophoric}     & Synthesize \texttt{CC(=O)N1c2ccc(N3CCNCC3)\allowbreak{}cc2{[}C@H{]}(Nc2ccccc2){[}C@@H{]}(C){[}C@@H{]}1C}   without using any pyrophoric or water-reactive reagents.                                                 \\
P9  &  \mbox{Pyrophoric}     & Find the shortest synthesis route for   \texttt{COc1cc2c(=O){[}nH{]}c(=O)n\allowbreak{}({[}C@@H{]}3O{[}C@H{]}(CO){[}C@H{]}4OC(C)(C)O{[}C@H{]}43)c2cc1OC},   avoiding pyrophoric substances.                         \\
P10 &  \mbox{Pyrophoric}     & Identify a synthesis route for   \texttt{Oc1ccc2c3c(ccc2c1)Cc1ccccc1OC3c1ccc(OCCN2CCCCC2)cc1} that does not use   pyrophoric substances.                                                               \\
P11 &  \mbox{Pyrophoric}     & Plan the synthesis of   \texttt{C{[}C@@H{]}(O)C{[}C@H{]}1OC{[}C@@H{]}\allowbreak{}(C2CCCCC2)N(c2cc(C\#CC(C)(C)C)sc2C(=O)O)C1=O}. Do   not use any pyrophoric substances.                                            \\
P12 &  \mbox{Pyrophoric}     & Find the shortest synthesis route for   \texttt{OC{[}C@H{]}1C{[}C@@H{]}(c2cnn3c(N{[}C@H{]}4CCc5ccccc54)\allowbreak{}ncnc23)C{[}C@@H{]}1O} that avoids   using any pyrophoric substances.                            \\
\midrule
S1  &  \mbox{User-Specified} & Find the shortest synthesis path for \texttt{C{[}C@@H{]}1CCCN1CCc1nnc2cc(Br)ccc2c1O},   but avoid using \texttt{C=C{[}Sn{]}(CCCC)(CCCC)CCCC} in your synthesis route.                                         \\
S2  &  \mbox{User-Specified} & Plan a synthesis route for   \texttt{CC(=O)NC{[}C@H{]}1CN(c2ccc3c(c2)\allowbreak{}CCCc2c(C(C)C)n{[}nH{]}c2-3)C(=O)O1}. Avoid using   phosgene (\texttt{O=C(Cl)Cl}) in your synthesis.                                      \\
S3  &  \mbox{User-Specified} & Find the shortest synthesis route for   \texttt{C{[}C@@H{]}1CNC(=O)c2cc3cc(OCCCN4CCCCC4)ccc3n21} that does not use trimethyl   borate (\texttt{COB(OC)OC}).                                                   \\
S4  &  \mbox{User-Specified} & Plan the synthesis of   \texttt{COCCCc1cc(CN(C(=O){[}C@H{]}2CNCC{[}C@@H{]}2c2ccc\allowbreak{}(OCCOc3c(Cl)cc(C)cc3Cl)cc2)C2CC2)cc(OCCOC)c1}   without using hexane (\texttt{CCCCCC}).                                       \\
S5  &  \mbox{User-Specified} & Identify a synthesis plan for   \texttt{COc1cc2c(=O){[}nH{]}c(=O)n({[}C@@H{]}3O{[}C@H{]}(CO){[}C@H{]}4OC(C)\allowbreak{}(C)O{[}C@H{]}43)c2cc1OC} that   does not use methanol (\texttt{CO}).                               \\
S6  &  \mbox{User-Specified} & Plan the synthesis of   \texttt{CC(C)c1ccc2c(c1)OC1(O)c3ccccc3C(=O)\allowbreak{}C21NC(=O)c1cc(-c2ccccc2)n{[}nH{]}1}. Do not use phenol (\texttt{Oc1ccccc1}).                                                             \\
S7  &  \mbox{User-Specified} & Synthesize \texttt{CCOC(=O)/C(N)=N/Nc1cc(Cl)ccc1{[}N+{]}(=O){[}O-{]}} without using   nitric acid (\texttt{O={[}N+{]}({[}O-{]})O}).                                                                           \\
S8  &  \mbox{User-Specified} & Find the shortest synthesis route for   \texttt{C\#CC1(O)C(C)=CC2(CC1(C)C(F)(F)F)OC(C)C(C)O2} that does not use toluene   (\texttt{Cc1ccccc1}).                                                               \\ 
\bottomrule
\end{longtable}
}

\clearpage
\section{\method System Instructions}
\label{app:prompts}

Here we present the system instructions for implementing the \evaluator in \methodmistral and \methodclaude.
Each set of system instructions was carefully engineered to elicit good instruction following from \mistral and \claude, respectively.

\subsection{\methodmistral System Instructions}

The following system instructions are used to implement the \evaluator in \methodmistral.
Values surrounded by curly braces \{\} are replaced with task-specific information.
Please note, reaction evaluation is broken into two separate instructions:
one to call tools and another to generate a score based on the results.

\paragraph{Evaluation Planning}

This set of instructions instructs \mistral to perform evaluation planning, generating the evaluation instructions.

\begin{promptbox}

You are an expert chemist evaluating a reaction. Your task is to use tools to determine if the user's provided reaction SMILES (reactants\textgreater{}\textgreater{}product) satisfies only the following constraint:

To help you evaluate the constraint, you may use any of the following tools:
\begin{itemize}
\item AIExpert(question): Asks a self-contained question of an AI chemistry expert to get information not available from the other tools or infer reaction conditions (e.g. catalysts, temperature, etc.).
\item Carcinogenicity(SMILES1, SMILES2, ...): returns a probabilistic score for each molecule indicating whether the provided SMILES represents a carcinogenic molecule. Can be called with one or more molecules; batching improves efficiency.
\item Pyrophoricity(SMILES1, SMILES2, ...): returns a score for each molecule indicating the Tanimoto similarity between the provided SMILES and its nearest neighbor on an authoritative list of pyrophoric substances (materials that spontaneously ignite in air or water). A score of 1.000 indicates the substance is on the list. Can be called with one or more molecules. The term "pyrophoric" includes water-reactive substances, as these may ignite in air due to the presence of water vapor.
\item Similarity(SMILES1, SMILES2): returns a similarity between exactly 2 molecules, represented by SMILES.
\end{itemize}

Create a plan of the actions you would take to evaluate this specific constraint for any user-provided reaction. Use as few actions as possible to evaluate the constraint, it is acceptable to use a single action. Explicitly identify actions taken using the following template:
\texttt{\\
\textasciigrave{}\textasciigrave{}\textasciigrave{} \\
ActionName(comma-separated action inputs) \\
ActionName(comma-separated action inputs) \\
... \\
\textasciigrave{}\textasciigrave{}\textasciigrave{}
}

Action names must be one of: AIExpert, Carcinogenicity, Pyrophoricity, Similarity.

In place of actual inputs, place descriptions of the values you would use. Be efficient and use only the tools necessary.
Then, reate a rubric for assigning a score of 1-5 where 5 is assigned if the constraint is completely satisfied and 1 is assigned if the constraint is completely violated.

Constraint: \{constraint\}

\end{promptbox}

\paragraph{Reaction Evaluation 1}

This set of instructions instructs \mistral to follow the evaluation instructions to select tools to evaluate a specific reaction.
It uses the messages from evaluation planning (including the generated evaluation instructions) as context.

\begin{promptbox}
\{evaluation planning context\}

User-provided reaction: \\
Reactants:
\{reactants\} \\
Product: \textasciigrave{}\{product\}\textasciigrave{}

When multiple reactants are present, they are separated by dots.

Fill in your plan with the specific inputs for this reaction. Format your actions using the following format, one per line, with inputs wrapped in quotes:
\texttt{\\
\textasciigrave{}\textasciigrave{}\textasciigrave{}\\
ActionName(ActionInput1, ActionInput2, ...) \\
ActionName(ActionInput1, ActionInput2, ...) \\
... \\
\textasciigrave{}\textasciigrave{}\textasciigrave{}
}

Do not include any other information.  Await the results.
\end{promptbox}

\paragraph{Reaction Evaluation 2}

This set of instructions instructs \mistral to generate a score between 1 and 5.  It uses the evaluation planning messages, the messages from Reaction Evaluation 1 and the tool outputs as context.

\begin{promptbox}
\{evaluation planning context\}

\{reaction evaluation context\}

\{tool outputs\}

Analyze these results and provide your final score. Use the following format:

Reasoning: (your analysis) \\
Answer: N

where N is between 1 and 5.
\end{promptbox}

\subsection{\methodclaude System Instructions}

The following system instructions are used to implement the \evaluator in \methodclaude.
Values surrounded by curly braces \{\} are replaced with task-specific information.
Please note, reaction evaluation is broken into two separate instructions:
one to call tools and another to generate a score based on the results.

\paragraph{Evaluation Planning}

This set of instructions instructs \claude to perform evaluation planning, generating the evaluation instructions.

\begin{promptbox}

You are an expert chemist assisting in retrosynthesis planning. You aim to determine if the user's provided reaction (reactants\textgreater\textgreater{}product) satisfies the following constraint:

\{constraint\}

Note that there may be reagents omitted from the user's provided reaction.

To help you answer the user question, you have access to these tools:
\begin{itemize}
    \item Carcinogenicity(SMILES1, SMILES2, ...): returns a probabilistic score for each molecule indicating whether the provided SMILES represents a carcinogenic molecule. Can be called with one or more molecules; batching improves efficiency.
    \item Pyrophoricity(SMILES1, SMILES2, ...): returns a score for each molecule indicating the Tanimoto similarity between the provided SMILES and its nearest neighbor on an authoritative list of pyrophoric substances (materials that spontaneously ignite in air or water). A score of 1.000 indicates the substance is on the list. Can be called with one or more molecules. The term "pyrophoric" includes water-reactive substances, as these may ignite in air due to the presence of water vapor.
    \item Similarity(SMILES1, SMILES2): returns a similarity between exactly 2 molecules, represented by SMILES.
    \item AIExpert(question): Asks a self-contained question of an AI chemistry expert, for instance, to infer reaction conditions or reagents not listed in the reaction, or to get information not available from the other tools. Include all relevant details, including reaction information and SMILES strings.
    \item Answer(value): Answer with a score of 1-5 based on only the constraint above is satisfied, where 5 is completely satisfied and 1 is not at all satisfied. Keep in mind these scores will be used to prioritize further planning.
\end{itemize}

Create a plan of the actions you would take to evaluate this specific constraint for any user-provided reaction. Explicitly identify actions taken using the following template:

Action: ActionName(ActionInput)

In place of actual inputs, place descriptions of the values you would use. Use only the tools necessary. Think step by step.

\end{promptbox}

\paragraph{Reaction Evaluation 1}

This set of instructions instructs \claude to follow the evaluation instructions to select tools to evaluate a specific reaction.

\begin{promptbox}    

You are an expert chemist assisting in retrosynthesis planning. You aim to determine if the user's provided reaction (reactants\textgreater\textgreater{}product) satisfies the following constraint:

\{constraint\}

To help you answer the user question, you have access to these tools:
\begin{itemize}
\item Carcinogenicity(SMILES1, SMILES2, ...): returns a probabilistic score for each molecule indicating whether the provided SMILES represents a carcinogenic molecule. Can be called with one or more molecules; batching improves efficiency.
\item Pyrophoricity(SMILES1, SMILES2, ...): returns a score for each molecule indicating the Tanimoto similarity between the provided SMILES and its nearest neighbor on an authoritative list of pyrophoric substances (materials that spontaneously ignite in air or water). A score of 1.000 indicates the substance is on the list. Can be called with one or more molecules. The term "pyrophoric" includes water-reactive substances, as these may ignite in air due to the presence of water vapor.
\item Similarity(SMILES1, SMILES2): returns a similarity between exactly 2 molecules, represented by SMILES.
\item Answer(value): Answer with a score of 1-5 based on only the constraint above is satisfied, where 5 is completely satisfied and 1 is not at all satisfied. Keep in mind these scores will be used to prioritize further planning.
\end{itemize}

In one response, specify all of the actions you would take to gather the necessary information to evaluate the user's reaction. Follow this plan:

\{evaluation instructions\}

Follow the plan step by step.
Replace all references to the AIExpert tool with your own expert chemistry knowledge on the user's reaction.
Keep in mind there may be multiple correct answers to each question (e.g. multiple ways to catalyze a reaction), and you should discuss every possibility in detail.
Reactants and products provided by the user should be considered required for the reaction, changing these would be considered a different reaction.
Provide accurate and valid SMILES representations for molecules in your answer.

For the tools, construct a single, unified code block wrapped in triple backticks at the end of your response, including specific inputs based on the user's provided reaction.
Ignore the Answer tool until you have results from your code block actions.
Wrap all SMILES in backticks.

\end{promptbox}

\paragraph{Reaction Evaluation 2}

This set of instructions instructs \claude to generate a score between 1 and 5.  It uses the messages from Reaction Evaluation 1 and the tool outputs as context.

\begin{promptbox}

\{reaction evaluation context\}

\{tool outputs\}

Given this information, please provide a final score for the reaction using \textasciigrave{}\textasciigrave{}\textasciigrave{}Answer(X)\textasciigrave{}\textasciigrave{}\textasciigrave{}, where X is your score between 1-5 on only this constraint:

\{constraint\}

Do not worry about any other constraints, as we may assess these separately.

\end{promptbox}

\section{Implementation Details for Reproducibility}
\label{app:implementation}

For the experiments in this paper,
we used the following hyperparameters and settings.
To control cost and runtime, we enforced a limit on the number of expansions and evaluations.
The search terminates after 500 expansions, 
following the established convention in multi-step retrosynthesis literature~\citep{chen2020retro, zhao_meea_2024, kim_self-improved_2021}.
Additionally, after 300 evaluations, all subsequent evaluations were replaced with the optimistic default score (i.e. 5 on the 1-5 scale).
To unify scaling, the raw values from {\meea}'s $\valuefn_{\scriptsize{\astar}}$ are min-max normalized to a $[0,1]$ scale at each selection step.
The {\meea} simulation step already included min-max scaling, so no change was required for $\valuefn_{MCTS}$.
Additionally, the evaluator scores $\score(\reaction)$
are each normalized to a $[0,1]$ scale, where 0 indicates complete constraint violation and 1 indicates complete constraint satisfaction.
For hyperparameters, we used $\lambda=2$ to ensure the constraint was followed and $K=5$ to control the number of candidate routes subject to evaluation.
Following the {\meea} paper, we used a UCB scaling term of 4.

The algorithm for \method is presented in Algorithm~\ref{alg:larc},
illustrating how \method uses both the \evaluator and \synthesizer to perform constrained retrosynthesis planning.
\method's \evaluator has two steps: evaluation planning (denoted $\evaluator\textsf{-plan}$ in Algorithm~\ref{alg:larc}) and reaction evaluation (denoted $\evaluator\textsf{-evaluate}$).
Both of these steps are implemented using the system instructions in Appendix~\ref{app:prompts}.
\method's \synthesizer adapts \meea's two step process: MCTS simulation (denoted $\synthesizer\textsf{-MCTS}$) to generate $K$ candidate routes, and \astar search (denoted $\synthesizer\textsf{-\astar}$) to select the best of the $K$ candidates.
\method's constrained adaptations of these steps
are given in Algorithms~\ref{alg:simulate} and~\ref{alg:select}, respectively.
\method's \synthesizer uses the same expansion function (denoted \textsc{Expand} in Algorithm~\ref{alg:larc}) as \meea~\cite{zhao_meea_2024}.
Please note, these algorithms assume $\valuefn_{\scriptsize\astar}$ and $S(r)$ are already normalized, per the implementation details above.
A full implementation of \method can be found at \url{https://github.com/ninglab/LARC}.

\newcommand{\candroutes}{\mbox{$\mathop{\mathcal{C}}\limits$}\xspace}

\newcommand{\expreactions}{\mbox{$\mathop{\mathcal{R}}\limits$}\xspace}

\newcommand{\expandfn}{\mbox{$\mathop{\textsc{Expand}}\limits$}\xspace}

\newcommand{\defaultscore}{\mbox{$\mathop{\score_\text{def}}\limits$}\xspace}

\begin{algorithm}[H]
\caption{\method}
\label{alg:larc}
\begin{algorithmic}[1]
\Require 
         Target molecule $p$;
         constraints $c$; 
         commercially available molecule set $B$; \Statex
         unconstrained MCTS value function $\valuefn_{MCTS}$; 
         unconstrained \astar value function $\valuefn_{\scriptsize\astar}$; \Statex
         expansion function \expandfn; \Statex
         expansion limit $N_{exp}$;
         evaluation limit $N_{eval}$; \Statex
         number of simulations $K$;
         constraint weight $\lambda$;
         default score $\defaultscore$ \Statex
\Ensure Synthetic route $\ancestors_*$
\Statex

\State $n_{exp} \gets 0$
\Comment{Initialize expansion count}
\State $T \gets$ Tree(nodes=\{p\}, edges=\{\})
\Comment{Initialize search tree}
\State $S \gets$ \{\}
\Comment{Initialize reaction evaluations} 
\Statex
\State $P \gets \evaluator\textsf{-plan}(c)$
\Comment{Evaluation planning generates evaluation instructions $P$}
\Statex

\While{$\ancestors_*$ is empty $\land$ $n_{exp} < N_{exp}$}
    \State $\candroutes \gets \synthesizer\textsf{-MCTS}(T, \valuefn_{MCTS}, \score, \defaultscore, \lambda, K)$
    \Comment{Simulate to get candidate set $\candroutes$} \\
    \ForAll{$(\molecule, \ancestors) \in \candroutes$}
        \ForAll{$\reaction \in \ancestors$}
            \If{$|S| < N_{eval} \land \reaction \notin \score$}
                \State $\score(\reaction) \gets \evaluator\textsf{-evaluate}(\reaction, P)$ 
                \Comment{Reaction evaluation}
            \ElsIf{$\reaction \notin \score$}
                \State $S(\reaction) \gets \defaultscore$ \Comment{Use default after evaluation limit}
            \EndIf
        \EndFor
    \EndFor
    \\
    \State ($\molecule_{exp}, \ancestors_{exp}) \gets \synthesizer\textsf{-\astar}(\candroutes, \valuefn_{\scriptsize\astar}, \score, \lambda)$ \Comment{Selection of ($\molecule_{exp}$, $\ancestors_{exp}$) from $\candroutes$}
    \\
    \State $\expreactions \gets \expandfn(T, \ancestors_{exp}, \molecule_{exp})$ \Comment{Expansion updates $T$ with set of reactions \expreactions}
    \State $n_{exp} \gets n_{exp} + 1$
    \Statex

    \ForAll{$\reaction \in \expreactions$}
        \If{$\ancestors_{exp} \oplus \reaction$ is a complete route from  $B$ to $p$} \Comment{Check for complete route}
            \State $\ancestors_* \gets \ancestors_{exp} \oplus \reaction$
        \EndIf
    \EndFor
\EndWhile \\

\State \Return $\ancestors_*$

\end{algorithmic}
\end{algorithm}

\begin{algorithm}[H]
\caption{\synthesizer-\textsf{MCTS}}
\label{alg:simulate}
\begin{algorithmic}[1]
\Require search tree $T$;
unconstrained value function (includes UCB) $\valuefn_{MCTS}$;
\Statex
\evaluator scores $\score$;
default score $\defaultscore$
constraint weight $\lambda$;
number of simulations $K$;
\Statex
\Ensure expansion candidates $\candroutes = \{(\molecule, \ancestors), ...\}$ of intermediate-route pairs
\Statex
\State $\candroutes \gets \{\}$
\State $\valuefn_{MCTS}^\prime(\molecule, \ancestors) := \valuefn_{MCTS}(m,R) \, + \, \lambda \sum\limits_{r\in R} \big( \score(r)\ \text{if } r\in \score\ \text{else } \defaultscore \big)$
\Comment{Equation~\ref{eqn:value}}
\Statex
\For{$k = 1$ to $K$}
    \State $\ancestors \gets \langle\;\rangle$; $\molecule \gets $ root$(T)$
    \Comment{start from target molecule and empty route}
    \State $A \gets$ reactions that produce $m$ in $T$
    \Statex
    \While{$|A| > 0$} \Comment{Simulate route from reactions in $T$}
        \State $\reaction^*, \molecule^* \gets \underset{\scriptsize\reaction \in A, \; \molecule \in \text{reactants}(\reaction)}{\arg\max} \valuefn_{MCTS}^\prime(\molecule, \ancestors \oplus \reaction)$ \Comment{select reaction}
        \State $\ancestors \gets \ancestors \oplus r^*$; \; $\molecule \gets \molecule^*$
        \Comment{update route}
        \State $A \gets$ reactions that produce intermediate $m$ in $T$
    \EndWhile
    \Statex
    \State $\candroutes \gets \candroutes \cup \{(\molecule, \ancestors)\}$
    \Comment{add candidate}
\EndFor
\Statex
\State \Return $\candroutes$
\end{algorithmic}
\end{algorithm}

\begin{algorithm}[H]
\caption{\synthesizer-\textsf{\astar}}
\label{alg:select}
\begin{algorithmic}[1]
\Require candidate routes \candroutes;
unconstrained value function $\valuefn_{\scriptsize\astar}$;
\Statex
\evaluator scores $\score$; constraint weight $\lambda$; 
\Statex
\Ensure intermediate molecule $\molecule_*$, route $\ancestors_*$
\Statex

\State $\candroutes \gets \{\}$ \Comment{Initialize empty candidate set}
\State $\valuefn_{\scriptsize\astar}^\prime(\molecule, \ancestors) := \valuefn_{\scriptsize\astar}(m,R) \, + \, \lambda \sum\limits_{r\in R}\score(r)$
\Comment{Equation~\ref{eqn:value}}
\Statex
\State $\molecule_*, \ancestors_* \gets \underset{\scriptsize{(\molecule, \ancestors)} \in \candroutes}{\arg\max} \; \valuefn_{\scriptsize\astar}^\prime(\molecule, \ancestors)$
\Comment{Select best candidate based on constrained \astar policy}
\State \Return $\molecule_*, \ancestors_*$
\end{algorithmic}
\end{algorithm}

\section{Additional Related Work}
\label{app:related}

\paragraph{LLM Agents for Chemistry}

LLM agents have recently shown great promise in chemistry applications.
\coscientist~\citep{boiko_autonomous_2023} and
\chemcrow~\citep{m_bran_augmenting_2024}, are notable examples, combining LLM reasoning with cheminformatics tools, web search, and robotic laboratory equipment to perform chemistry tasks, including retrosynthesis planning.
However, neither of these methods are well-equipped for constrained retrosynthesis planning.
\coscientist relies on general-purpose LLMs for retrosynthesis planning, 
which lack accuracy on chemical reaction tasks compared to models with specialized training~\citep{yu_llasmol_2024}.
\chemcrow generates synthetic routes using an external tool for unconstrained retrosynthesis planning,
making it incapable of incorporating constraints directly into the planning process.
\cactus~\citep{mcnaughton_cactus_2024} and \chemtoolagent~\citep{yu_chemtoolagent_2025} are similar to \chemcrow, but with different focuses.
\cactus focuses only on molecule property prediction with in silico tools,
and \chemtoolagent focuses on analyzing the impact of tool use across various chemistry tasks.
\cite{gottweis_towards_2025} introduces an LLM agent co-scientist similar to \coscientist,
but designed for broad applicability across multiple scientific disciplines,
not specifically for chemistry or retrosynthesis planning.
\liddia~\citep{averly_liddia_2025} is an LLM agent for in silico drug discovery that automates generation, screening, and optimization of molecules, but it does not consider retrosynthesis planning.
While these efforts demonstrate the promise of LLM agents in chemistry applications,
none of the existing LLM agents can perform constrained retrosynthesis planning.

\section{Additional Experimental Settings}
\label{app:settings}

\paragraph{Constrained retrosynthesis planning by LLMs} We use general-purpose LLMs, 
including \claude~\citep{anthropic_claude_2024}, \gpt~\citep{openai_gpt-4o_2024}, \deepseek~\citep{deepseek_2025}, and \mistral~\citep{mistral_nemo_2024} as the baselines for constrained retrosynthesis planning. 
\claude and \gpt are representative state-of-the-art closed-source LLMs,
which have both shown promising results on chemistry knowledge benchmarks~\citep{wang_mmlu-pro_2024, rein_gpqa_2024, yu_chemtoolagent_2025}.
\deepseek serves as a representative of open-source reasoning LLMs, leveraging an internal chain of thought~\citep{wei_chain_2022} process to improve its responses.
\mistral serves as a representative of small, open-source LLMs, selected due to its instruction-following capabilities~\citep{mistral_nemo_2024}. 
Note that baselines do not include models specifically designed for unconstrained retrosynthesis planning, 
such as MEEA. 
This is because the benchmark is constructed by selecting the tasks for which  
MEEA can generate valid routes but violate the constraints. 
As MEEA is the state of the art for unconstrained retrosynthesis planning, we believe other unconstrained 
retrosynthesis planning methods~\citep{chen2020retro, kim_self-improved_2021,genheden_aizynthfinder_2020,segler_planning_2018} will not succeed on the benchmark data. 

\paragraph{Constrained retrosynthesis planning by human experts}
We also compare \method against a human expert in retrosynthesis planning, denoted as \Expert.
\Expert is an experienced synthetic chemist with a doctoral degree and over 17 years of experience in retrosynthesis planning.
For each task, \Expert was provided with clear instructions on the constraint and the desired chemical product.
\Expert had full access to necessary resources and reference materials, including lists of purchasable materials~\citep{chen2020retro}, 
carcinogenic chemicals, and pyrophoric chemicals, as well as external and online 
references such as Reaxys~\citep{reaxys} and SciFinder~\citep{scifinder}. 
However, 
\Expert was not permitted to use computer-aided multi-step retrosynthesis planning tools, such as AIZynthFinder~\citep{genheden_aizynthfinder_2020} or \meea~\citep{zhao_meea_2024}. 
To ensure both the quality and efficiency of human planning, \Expert was instructed to complete each task with careful attention 
to detail while minimizing planning time. 
To support sustained performance and maintain high-quality planning, \Expert was encouraged to take a 15-minute break between tasks.

\section{Additional Results}
\label{app:results}

\begin{figure*}[htb]
\centering
\includegraphics[width=\linewidth]{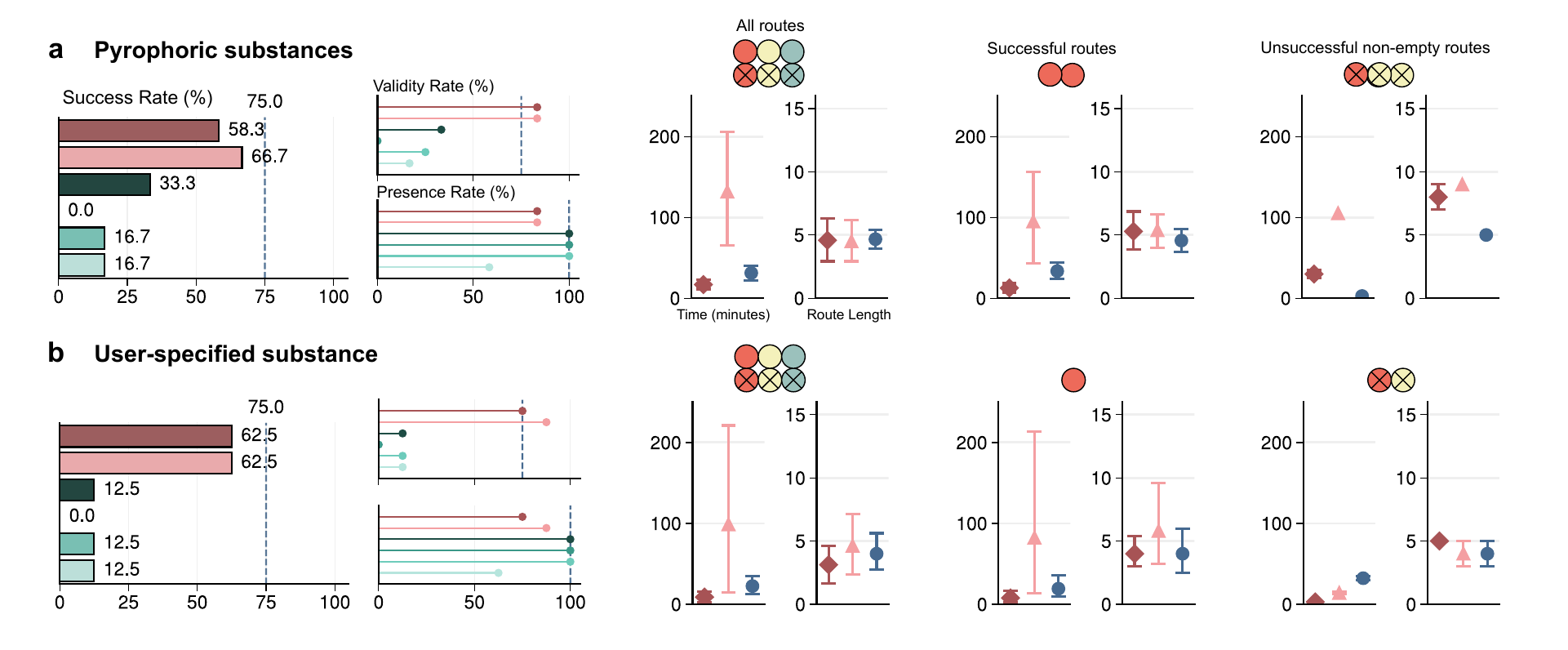}
\vspace{-20pt}
\caption{\textbf{a,b} success rate, validity rate, presence rate,
planning time, and route length for the benchmark tasks, organized by constraints 
for avoiding
 pyrophoric substances,
and a user-specified substance, respectively.
The symbols and colors in this figure follow the same conventions as Figure~\ref{fig:results}.
}
\label{fig:results:full}
\end{figure*}

\subsection{Pyrophoricity-constrained retrosynthesis planning}
\label{app:results:pyro}

Figure~\ref{fig:results:full}\textbf{e} presents the results for retrosynthesis planning tasks 
that are constrained to avoid pyrophoric substances.
\methodmistral and \methodclaude continue to achieve high success rates of 58.3\% and 66.7\%, respectively.  
\methodclaude moderately underperforms \Expert, who secures a success rate of 75.0\%. 
Meanwhile, \methodmistral is significantly faster, with 12.58 minutes on average to generate one successful route, 
compared to \Expert (33.48 minutes) (p=0.020, two-sided two-sample t-test)
and \methodclaude (94.81 minutes) (p=0.026).
This suggests that, for retrosynthesis planning avoiding pyrophoricity, 
\method is still behind human experts, but could be a decent option with satisfactory/acceptable 
success rates, considering its efficiency and automation.
\method can also serve as an alternative to human experts, and provide additional, different solutions, 
which human experts can further select, utilize, or optimize based on their domain knowledge.  

The LLM baselines fall short on these tasks, with the best LLM baseline, 
\claude, achieving only 33.3\% success rate. %
Pyrophoric substances can vary greatly in their composition and the underlying mechanism of pyrophoricity~\citep{gibson_jack_r_handbook_1969},
demanding a nuanced understanding of chemical reactivity.
These results demonstrate \method's
ability and efficiency to avoid a broad set of dangerous materials in its retrosynthesis planning
compared to general-purpose LLMs. 

\subsection{User-specified constrained retrosynthesis planning}
\label{app:results:user}

As Figure~\ref{fig:results:full}\textbf{f} hows, 
\methodclaude and \methodmistral achieve the same success rate of 62.5\% 
when planning 8 routes that must each avoid a single, user-specified substance.
LLMs still struggle, with a very low success rate of 12.5\%, due to the lack of 
ability to generate valid synthetic routes.
On the other hand, \method 
generalizes well to the highly diverse constraint types, 
fulfilling a key need in the dynamic context of retrosynthesis planning.
\Expert achieves a higher success rate of 75.0\% with an average %
22.47 minutes on generating one route. 
\methodmistral is still faster on average than \Expert (8.82 minutes), 
though the difference is not very significant (p=0.089, two-sided two-sample t-test).

\subsection{Case Studies}
\label{app:results:case}

\begin{figure*}[htbp]
\includegraphics[width=\linewidth]{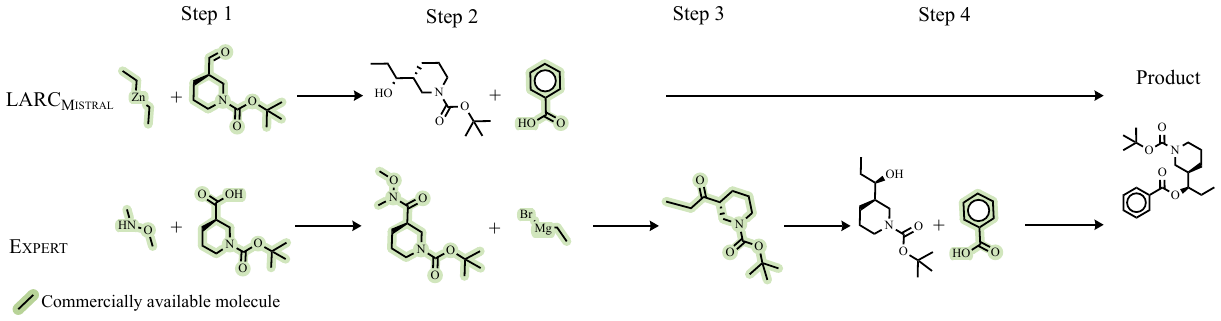}
\caption{Synthesis route comparison of \methodmistral against \Expert tert-butyl (R)-3-((R)-1-\allowbreak(benzoyloxy)propyl)\allowbreak-piperidine-1-\allowbreak{}carboxylate.}
\label{fig:case:better}
\end{figure*}

\subsubsection{\methodmistral can generate better routes than human experts}
\label{sec:results:case:better}
Figure~\ref{fig:case:better} 
compares routes from \methodmistral and \Expert for the synthesis of 
tert-butyl (R)-3-((R)-1-\allowbreak(benzoyloxy)propyl)\allowbreak-piperidine-1-carboxylate.
\methodmistral proposed a very concise two-step reaction to the product 
using commercially available reagents: 
diethyl zinc, tert-butyl (S)-3-formylpiperidine-1-carboxylate, and benzoic acid.
Its Step 1 involves a stereoselective, nucleophilic addition of an ethyl group to tert-butyl (S)-3-formylpiperidine-1-carboxylate using the diethylzinc reagent.
This results in an alcohol that can be converted to the product (an ester) using benzoic acid in Step 2.

In contrast, \Expert proposed a four-step reaction to the same product.
Step 1 involves conversion of commercially available (S)-1-\allowbreak(tert-butoxycarbonyl)piperidine-3-\allowbreak{}carboxylic acid to the Weinreb amide intermediate,
tert-butyl (S)-3-(methoxy(methyl)carbamoyl)piperidine-1-carboxylate.
This step was unnecessary, as its product is also commercially available.
Step 2 is the Grignard addition of an ethyl group to obtain tert-butyl (S)-3-propionylpiperidine-1-carboxylate.
Stereoselective reduction of the ketone in Step 3 results in the alcohol intermediate obtained in {\methodmistral}’s Step 1.
A final ester formation, similar to that proposed by \methodmistral, yields the product.

One potential reason for \Expert's longer route is 
the \Expert's inherent bias towards the stability of some functional groups and 
their commercial availability.
For example, it is a generally accepted notion that aldehydes are not very stable~\citep{ouellette_aldehydes_2014},
so chemists tend to synthesize aldehydes and use them as needed.
In most cases, an aldehyde equivalent such as the Weinreb amide, which was proposed in Step 1 by \Expert,
is used instead, followed by a reduction step
when the product is an alcohol.
Thus, as the Weinreb amide is stable and easily synthesizable, 
\Expert may tend to propose a synthesis of this intermediate in the synthetic route.
This may be another reason why \Expert proposed this route,
relying on their familiarity with this synthesis strategy
rather than referring to the commercial availability of proposed intermediates.
In contrast, \methodmistral benefits from \synthesizer's thorough checks for commercial availability while simultaneously relying on feedback from the \evaluator to ensure the constraint is satisfied.

\subsection{Study on tooling in \method}
\label{app:results:tool}

\evaluator's interpretation of results from different tools also significantly impacts \method performance.
For example, 
in pyrophoricity-constrained retrosynthesis planning, \methodclaude performs better than \methodmistral 
(success rate of 66.7\% vs 58.3\%).
One reason for this lies in how \claude- and \mistral-based \evaluator interprets and leverages evaluation results from 
the pyrophoricity predictor. 
Although with similar instructions, 
\claude considers a high value (e.g., close to 1.0) from the predictor 
as an indicator of high pyrophoricity, and thus, a low $\score(\reaction)$. 
This allows {\synthesizer} to explore more extensively to avoid pyrophoricity, 
and thus, a high success rate in the generated routes.    
However, \methodmistral could interpret a relatively low value (e.g., 0.333) for high pyrophoricity, thus discouraging the exploration of routes that satisfy the constraint, resulting in more failed routes.

\section{Additional Discussion}
\label{app:discussion}

Constrained retrosynthesis planning is highly challenging in the practice of synthetic chemistry, 
as it requires finding synthetic routes that not only lead to the target molecule 
but also satisfy additional, often complex, user-specified requirements. 
These constraints, such as limiting the number of steps, avoiding specific reagents or reaction types, 
or adhering to cost, safety, or environmental guidelines, 
can drastically narrow and fragment the feasible search space. 
Moreover, verifying constraint satisfaction is non-trivial: 
it may involve detailed reaction feasibility assessments, availability checks for intermediates, 
or compliance with regulatory and safety standards. 
Computational tools for such verifications can be resource-intensive to run, less reliable, or even unavailable,  
while manual verification can be very slow, biased, and inconsistent --
it is not uncommon that chemists disagree with each other in retrosynthesis planning~\citep{takaoka_development_2003, lajiness_assessment_2004}.

\end{document}